\documentclass{article}
\usepackage{spconf,amsmath,graphicx}
\usepackage{subcaption}
\usepackage{algorithm}
\usepackage{algorithmic}
\usepackage{amsmath,amsfonts,amssymb,amsthm,bm}  
\usepackage{hyperref}
\usepackage{tikz}

\newcommand{\hide}[1]{} %To hide large blocks of code without using % symbols

\newcommand\Ebb{\ensuremath{\mathbb{E}}}

\newcommand\Rbb{\ensuremath{{\mathbb{R}}}}

\newcommand\Oc{\ensuremath{{\mathcal{O}}}}

\usepackage{tikz}

\newcommand{\sign}[1]{{\ensuremath{{\rm Sign}\left(#1\right)}}}

\newtheorem{assumption}{Assumption}

\newtheorem{theorem}{Theorem}

\title{FedLion: Faster Adaptive Federated Optimization \\ with Fewer Communication}
\name{Zhiwei Tang, Tsung-Hui Chang}
\address{School of Science \& Engineering, The Chinese University of Hong Kong, Shenzhen, China\\
Shenzhen Research Institute of Big Data, Shenzhen, China}
\begin{document}
%\ninept
%
\maketitle
\begin{abstract}
In Federated Learning (FL), a framework to train machine learning models across distributed data, well-known algorithms like FedAvg tend to have slow convergence rates, resulting in high communication costs during training.
To address this challenge, we introduce FedLion, an adaptive federated optimization algorithm that seamlessly incorporates key elements from the recently proposed centralized adaptive algorithm, Lion \cite{chen2023symbolic}, into the FL framework. Through comprehensive evaluations on two widely adopted FL benchmarks, we demonstrate that FedLion outperforms previous state-of-the-art adaptive algorithms, including FAFED \cite{wu2023faster} and FedDA \cite{in2022accelerated}. Moreover, thanks to the use of signed gradients in local training, FedLion substantially reduces data transmission requirements during uplink communication when compared to existing adaptive algorithms, further reducing communication costs. Last but not least, this work also includes a novel theoretical analysis, showcasing that FedLion attains faster convergence rate than established FL algorithms like FedAvg. 
\end{abstract}
\vspace{-0.1cm} 
\section{Introduction}
\label{sec:intro}\vspace{-0.1cm} 

\subsection{Federated Learning}
In this study, we focus on Federated Learning (FL), a privacy-preserving distributed learning paradigm that has garnered increasing interest in recent years. FL entails a scenario where numerous clients collaboratively train a machine learning model while safeguarding their local data by not sharing it with each other or with a central server. Instead, the clients engage with a central server solely for the exchange of model parameters or gradients. Our primary focus centers on the scenario featuring one centralized parameter server and $N$ clients, as established in prior work \cite{mcmahan2017communication}. Specifically, we aim to address the following distributed optimization problem:

\vspace{-0.5cm} 
\begin{align}
\label{p:problem}
\min_{x\in\Rbb^d} f(x) = \frac{1}{N}\sum_{i=1}^N f_i(x).\vspace{-0.6cm} 
\end{align}
where each client's local objective function, denoted as $f_i(\cdot)$, is generated from the dataset owned by that respective client.

When designing distributed algorithms to tackle \eqref{p:problem}, an important facet to consider is communication efficiency, since a substantial number of clients need to frequently transmit their local models/gradients to the central parameter server. As the most well-known FL algorithm, the Federated Averaging (FedAvg) algorithm \cite{mcmahan2017communication} employs multiple local Stochastic Gradient Descent (SGD) updates with periodic communications to curtail communication costs. Notwithstanding the communication reduction achieved through local SGD updates, FedAvg still imposes significant communication costs due to its relatively slow convergence. Therefore, it is crucial to design new algorithms that can achieve faster convergence than FedAvg, thereby further reducing the communication cost.

\vspace{-0.2cm} 
\subsection{Adaptive Optimization for Federated Learning}\vspace{-0.1cm} 
In the field of centralized neural network training, there exists a class of optimization algorithms known as adaptive algorithms. These algorithms leverage historical gradient information to enhance subsequent optimization iterations. Prominent examples include Stochastic Gradient Descent (SGD) with momentum, Adam \cite{kingma2014adam}, and the more recently introduced Lion \cite{chen2023symbolic}. Notably, Lion stands out as a state-of-the-art adaptive algorithm, consistently demonstrating superior convergence performance compared to Adam across a spectrum of centralized machine learning tasks.

Drawing inspiration from these centralized adaptive optimization algorithms, some previous research has sought to incorporate adaptive optimization techniques into federated learning. The aim is to accelerate convergence, with notable examples including works such as \cite{reddi2020adaptive, wu2023faster, in2022accelerated, das2022faster, liu2020accelerating}. Currently, the state-of-the-art algorithms in this field are FAFED \cite{wu2023faster} and FedDA \cite{in2022accelerated}, both of which demonstrate faster convergence than FedAvg \cite{mcmahan2017communication} and earlier adaptive federated learning algorithms like \cite{karimireddy2020scaffold, reddi2020adaptive, liu2020accelerating}.

However, a common drawback in all existing algorithms is that they typically require transmitting two or three times more bits than the standard FedAvg algorithm during the uplink transmission. This is because, aside from the local models/gradients, they also require transmitting auxiliary variables like local momentum or control variables, which necessitate an equal amount of bits as the local model.

In this paper, we introduce a new adaptive federated optimization algorithm named FedLion. FedLion adapts the cutting-edge centralized adaptive algorithm Lion \cite{chen2023symbolic} to the federated learning setting. We show that, both empirically and theoretically, FedLion achieves faster convergence than previous state-of-the-art algorithms like FAFED \cite{wu2023faster} and FedDA \cite{in2022accelerated}, all while requiring only a slightly more bits transmission compared to the FedAvg algorithm during the uplink transmission.
\vspace{-0.2cm} 

\section{Adaptive Federated Optimization with Lion}
\label{sec:method}\vspace{-0.2cm} 

\subsection{Periodic Averaging for Models and Momentums}\label{sec:algo}\vspace{-0.2cm} 
To design an adaptive FL algorithm, prior research efforts \cite{liu2020accelerating, wu2023faster} have introduced a straightforward yet effective approach for extending centralized adaptive algorithms like SGD with Momentum (SGDwM) and Adam into the federated learning paradigm: Periodically averaging both local models and local auxiliary variables (e.g., momentums). Notably, this approach aligns the FL algorithm with the same set of hyperparameters as its centralized counterpart, rendering it more practically convenient compared to other adaptive federated learning algorithms presented in prior works. For example, the algorithms in \cite{reddi2020adaptive, karimireddy2020scaffold, in2022accelerated} require meticulous tuning to strike a balance between local and global learning rates.

In accordance with this strategy, we introduce FedLion in Algorithm \ref{alg:FedLion}, which seamlessly integrates the newly established state-of-the-art centralized adaptive algorithm Lion in  \cite{chen2023symbolic} into the FL paradigm.

\begin{algorithm}[htbp]
  \small
    \begin{algorithmic}[1]
      \REQUIRE Total communication rounds $T$, number of local steps $E$, learning rate $\gamma$, momentum coefficient $(\beta_1,\beta_2)$, batchsize $B$, number of participant clients per round $n$.
      \STATE Initialize global model $x_0$ randomly and global momentum $m_0$ with zero elements.  
    \FOR{$t=1$ to $T$}
  
    \STATE \textbf{On Clients:}
    \STATE Randomly and uniformly sample $n$ clients ($n<N$).
    \FOR{$i=1$ to $n$}
    \STATE $x_{t-1,0}^i=x_{t-1}$ (Receive global model)
    \STATE $m_{t-1,0}^i=m_{t-1}$ (Receive global momentum)
    \FOR{$s=1$ to $E$}
    \STATE Compute gradient $g_{t-1,s}^{i}$ with $B$ minibatch samples.
    \STATE $h_{t-1,s}^i = \beta_1m_{t-1,s-1}^i+(1-\beta_1)g_{t-1,s}^{i}$
    \STATE $h_{t-1,s}^i=\sign{h_{t-1,s}^i}$
    \STATE $x_{t-1,s}^i =x_{t-1,s-1}^i - \gamma h_{t-1,s}^i$
    \STATE $m_{t-1,s}^i = \beta_2m_{t-1,s-1}^i+(1-\beta_2)g_{t-1,s}^{i}$
    \ENDFOR
    \STATE $\Delta_{t-1}^{i} = \left(x_{t-1}-x_{t-1,E}^i\right)/{\gamma}$ (Integer-valued)
    \STATE Send $\Delta_{t-1}^{i}$ and $m_{t-1,E}^i$ to the server.
    \ENDFOR
  
    \STATE \textbf{On Server:}
  
        \STATE  $x_t = x_{t-1}-  \frac{\gamma}{n}\sum_{i=1}^n  \Delta_{t-1}^{i}$ (Update global model)
        \STATE $m_t=\frac{1}{n}\sum_{i=1}^n m_{t-1,E}^i$ (Update global momentum)
    \ENDFOR
    \end{algorithmic}
    \caption{FedLion}
    \label{alg:FedLion}
    \end{algorithm} 

    We emphasize that Algorithm \ref{alg:FedLion} exhibits clear superiority in uplink communication efficiency over existing adaptive algorithms, such as those outlined in \cite{wu2023faster, karimireddy2020scaffold, liu2020accelerating}. An observation from Step 15 of Algorithm \ref{alg:FedLion} reveals that: 
    
    \vspace{-0.3cm} $$\Delta_{t-1}^{i} = \left(x_{t-1}-x_{t-1,E}^i\right)/{\gamma}=\sum_{s=1}^E h_{t-1,s}^i.$$ \vspace{-0.3cm} 
    
    Thanks to the sign operation, this implies that $\Delta_{t-1}^{i}$ takes on integer values within the range $[-E,E]$, necessitating no more than $\log(2E+1)$ bits for representation. Consequently, this approach results in notable bit savings during communication, as each client is only required to transmit an integer-valued vector and a full-precision vector, which requires only slightly more transmitted bits compared to the  FedAvg \cite{mcmahan2017communication}. This stands in contrast to the existing approaches like \cite{wu2023faster, in2022accelerated} that demand the transmission of two or three full-precision vectors.

\vspace{-0.2cm} 
\subsection{Convergence Analysis}
We delve into the convergence analysis of Algorithm \ref{alg:FedLion}. It is important to emphasize that, although Algorithm \ref{alg:FedLion} is a conceptually simple extension of Lion \cite{chen2023symbolic}, demonstrating its convergence is a challenging task, and there is even no convergence analysis provided for centralized Lion in \cite{chen2023symbolic}. To examine its convergence behavior, we introduce innovative theoretical tools alongside a novel bounded heterogeneity assumption. We initiate our analysis by presenting the essential assumptions required for our convergence study. For any vector $v\in\Rbb^d$, we denote $v(j)$ as the $j$-th element of $v$.

\begin{assumption} \label{asp:common}
\ 	\begin{enumerate}\vspace{-0.2cm} 
		\item[A.1 ] The minibatch gradient is unbiased and has bounded variance, i.e., $ \Ebb[g_i(x)]=\nabla f_i(x)$ and $$\Ebb[|g_i(x)(j)-\nabla f_i(x)(j)|^2]\leq \sigma_j^2,\ \forall j.$$\vspace{-0.7cm} 

		\item[A.2] Each $f_i$ is smooth, i.e., for any $x,y\in\Rbb^d$, there exists some non-negative constants $L_1,\ldots,L_d$, such that
			$$|\nabla f(y)(j)-\nabla f(x)(j)|\leq L_j,\ \forall j.$$\vspace{-0.7cm} 

		\item[A.3]$f$ is lower bounded, i.e., there exists some constant $f^*$ such that
			$f(x)\geq f^*,\forall x\in\Rbb^d.$\vspace{-0.3cm} 
   
		\item[A.4]  There exists a constant $\alpha\leq \frac{1}{3}$, such that $$\|\nabla f(x) - \nabla f_i(x)\|_1\leq \alpha \|\nabla f(x)\|_1, \ \forall i,x.$$\vspace{-0.5cm} 
	\end{enumerate}
	\end{assumption}\vspace{-0.3cm} 

  Assumptions A.1-A.3 are the common assumptions employed in prior studies, such as \cite{bernstein2018signsgd} and \cite{safaryan2021stochastic}, particularly when analyzing sign-based optimization algorithms. To prove the convergence Algorithm \ref{alg:FedLion}, we introduce a new assumption, A.4 in this work, with the specific aim of constraining the heterogeneity within local objective functions.

It is worth noting that A.4 represents a mildly stronger constraint when compared to existing bounded heterogeneity assumptions like A.5 used in \cite{9556559}, which takes the form: 
  $$\|\nabla f(x) - \nabla f_i(x)\|_2\leq  \delta, \ \forall i,x.$$ This suggests that our proposed algorithm is particularly suitable in scenarios with moderate levels of heterogeneity. Moreover, the trade-off towards reduced heterogeneity leads to an enhanced convergence rate than existing algorithms like FedAvg, as substantiated by the results in Theorem \ref{thm:convergence}. Although Algorithm \ref{alg:FedLion}  is limited to the setting characterized by A.4, empirical observations highlight its exceptional performance on the two most prevalent federated benchmarks and thus underline its practical utility.

\vspace{-0.1cm}
\begin{theorem}
  \label{thm:convergence} Suppose that A.1-A.4 in Assumption \ref{asp:common} hold. 
  Denote that $\bar x_{t,s} = \frac{1}{N}\sum_{i=1}^N x_{t,s}^i$, $\bar L = \sum_{j=1}^d L_j$ and $\bar \sigma = \sum_{j=1}^d \sigma_j$. If we set $\gamma=\frac{1}{\sqrt{T}}$, $\beta_1 = 1-\frac{1}{\sqrt{T}}$ and $\beta_2=1-\frac{1}{{T}}$ for Algorithm \ref{alg:FedLion}, then there exists three positive constants $C_1$, $C_2$ and $C_3$ such that
  \begin{align}
   & \frac{1}{TE}\sum_{t=1,...T}\sum_{s=1,...,E}\Ebb\left[\|\nabla f(\bar x_{t-1,s-1})\|_1\right]\notag\\ \label{p:convergence} \leq & \frac{1}{1-3\alpha}\left(\frac{f(x_0)-f^*}{E\sqrt{T}}+\frac{C_1\bar L}{\sqrt{T}}+\frac{C_2\bar \sigma}{\sqrt{nT}}+\frac{C_3\bar L E}{\sqrt{T}}\right).
  \end{align}\vspace{-0.2cm}
\end{theorem}

As observed, the optimal choice for the parameter $E$ depends on the geometric characteristics of the problem. Specifically, it involves balancing the contributions of the first and last terms in \eqref{p:convergence}. For instance, by setting $E$ to a value like $\left({{f(x_0)-f^*}/{C_3\bar L}}\right)^{\frac{1}{2}}=\Oc(1)$, we achieve a convergence rate of $\Oc({T^{-\frac{1}{2}}})$ for the $\ell_1$ norm of the function gradient. Conversely, if we opt for $E=\Oc({T^{-\frac{1}{4}}})$, a typical choice in the FedAvg algorithm \cite{yu2019parallel}, we attain the convergence rate depicted in \eqref{p:convergence2}.

\vspace{-0.3cm}
\begin{align}\label{p:convergence2}
 \left(\frac{1}{TE}\sum_{t,s}\Ebb\left[\|\nabla f(\bar x_{t-1,s-1})\|_1\right]\right)^2\leq \Oc\left(T^{-\frac{1}{2}}\right).
\end{align}\vspace{-0.3cm}

Examining the convergence bound in \eqref{p:convergence2}, we observe that the squared $\ell_1$ norm of the function gradient converges at a rate of $\Oc\left(T^{-\frac{1}{2}}\right)$. We note that this rate can surpass the $\Oc\left((nT)^{-\frac{1}{2}}\right)$ convergence rate of FedAvg (Corollary 1 in \cite{yu2019parallel}) in certain scenarios. Specifically, we recall that $ \|v\|_2\leq \|v\|_1$ holds for any vector $v\in\Rbb^d$. Moreover, if the gradient is dense enough, i.e., $\frac{\|v\|_1}{\|v\|_2}\gg n^{\frac{1}{4}}$, our proposed algorithm can be much faster than the rate $\Oc\left((nT)^{-\frac{1}{2}}\right)$ with squared $\ell_2$ norm as the convergence metric. In Section \ref{sec:grad_dense}, we empirically show that this is indeed the case that $\frac{\|v\|_1}{\|v\|_2}\gg n^{\frac{1}{4}}$, which justifies the faster convergence against other algorithms in empirical evaluations.

\section{Experiments}
\label{sec:exp}\vspace{-0.2cm} 

\subsection{Settings}

In this section, we assess the performance of FedLion on two widely recognized benchmarks commonly employed in FL \cite{reddi2020adaptive}: EMNIST and CIFAR-10.
For EMNIST, our evaluation encompasses a total of 3,579 clients. During each communication round, 100 clients are uniformly selected to transmit their compressed gradients.
The CIFAR-10 features 100 clients, with training samples distributed among them. Each client possesses a multinomial distribution over labels derived from a symmetric Dirichlet distribution with a parameter of 1. In every communication round, 10 out of the 100 clients are uniformly sampled for participation. For network architecture, we employ a 2-layer CNN for the EMNIST dataset and a ResNet18 with group normalization for CIFAR-10.

Our comparative analysis involves the baseline Federated Averaging (FedAvg) \cite{mcmahan2017communication} as well as three other existing adaptive FL algorithms: MFL \cite{liu2020accelerating}, FedDA \cite{in2022accelerated}, and FAFED \cite{wu2023faster}. Concerning hyperparameters, we meticulously select the optimal learning rates for FedAvg, MFL, and FAFED from the set $\{0.1, 0.01, 0.001\}$, respectively. In the case of MFL and FAFED, we retain their default hyperparameters, namely 0.9 and 0.99 for the first-order and second-order momentum coefficients. For FedDA \cite{in2022accelerated}, we adopt the FedDA+SGDwM version and employ the same hyperparameters used in its original experiment setup. We set $\gamma=0.001$, $\beta_1=0.9$, and $\beta_2=0.99$ for FedLion in all experiments. To facilitate reproducibility, we provide the code for replicating these experiments in this repository\footnote{\url{https://github.com/TZW1998/FedLion}}.

\vspace{-0.2cm} 
\subsection{Results}
\begin{figure}[t]
	\centering
 \vspace{-0.cm}
 \begin{subfigure}[b]{0.45\textwidth}
	\centering
	\includegraphics[width=\textwidth]{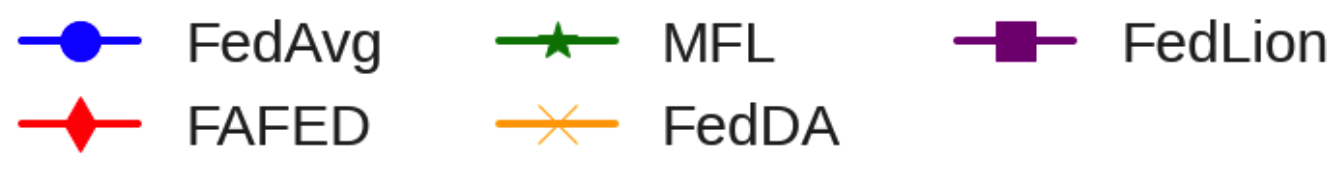}
	\label{}
\end{subfigure}
\vspace{-0.1cm}
	\begin{subfigure}[b]{0.15\textwidth}
	\centering
	\includegraphics[width=\textwidth]{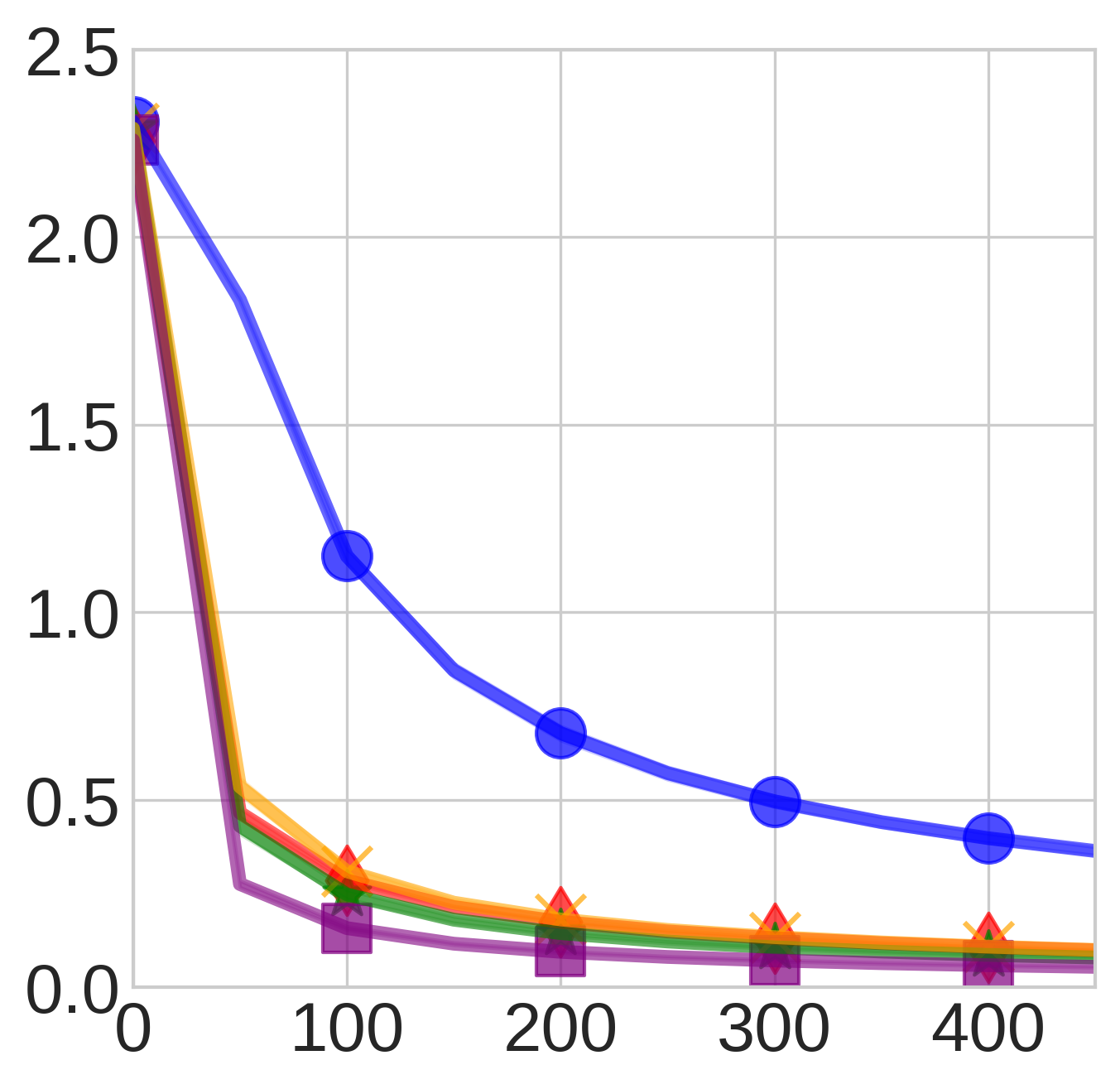}
	\label{}
\end{subfigure}\vspace{-0.1cm}
\begin{subfigure}[b]{0.15\textwidth}
	\centering
	\includegraphics[width=\textwidth]{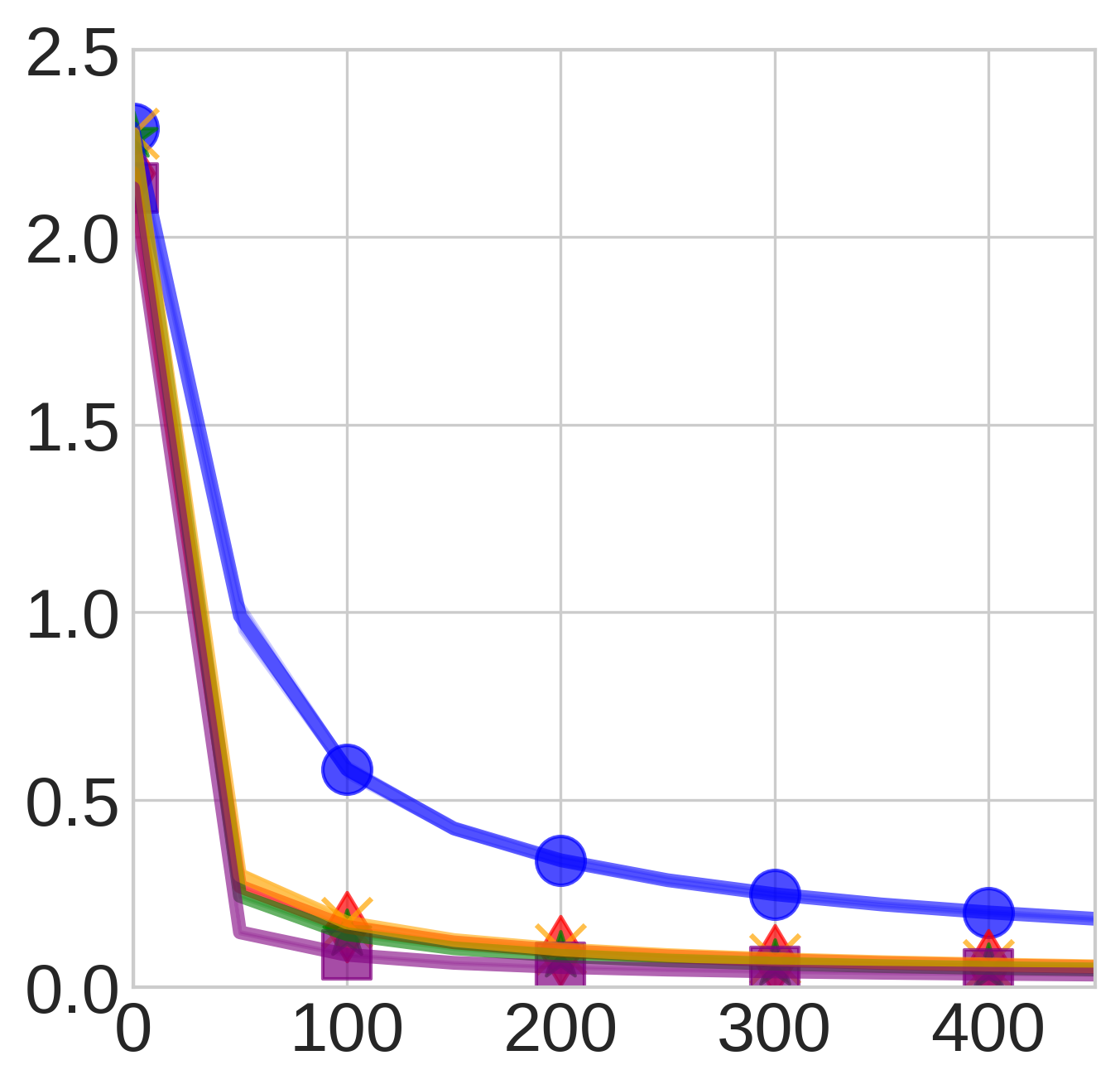}
	\label{}
\end{subfigure}\vspace{-0.1cm}
\begin{subfigure}[b]{0.15\textwidth}
	\centering
	\includegraphics[width=\textwidth]{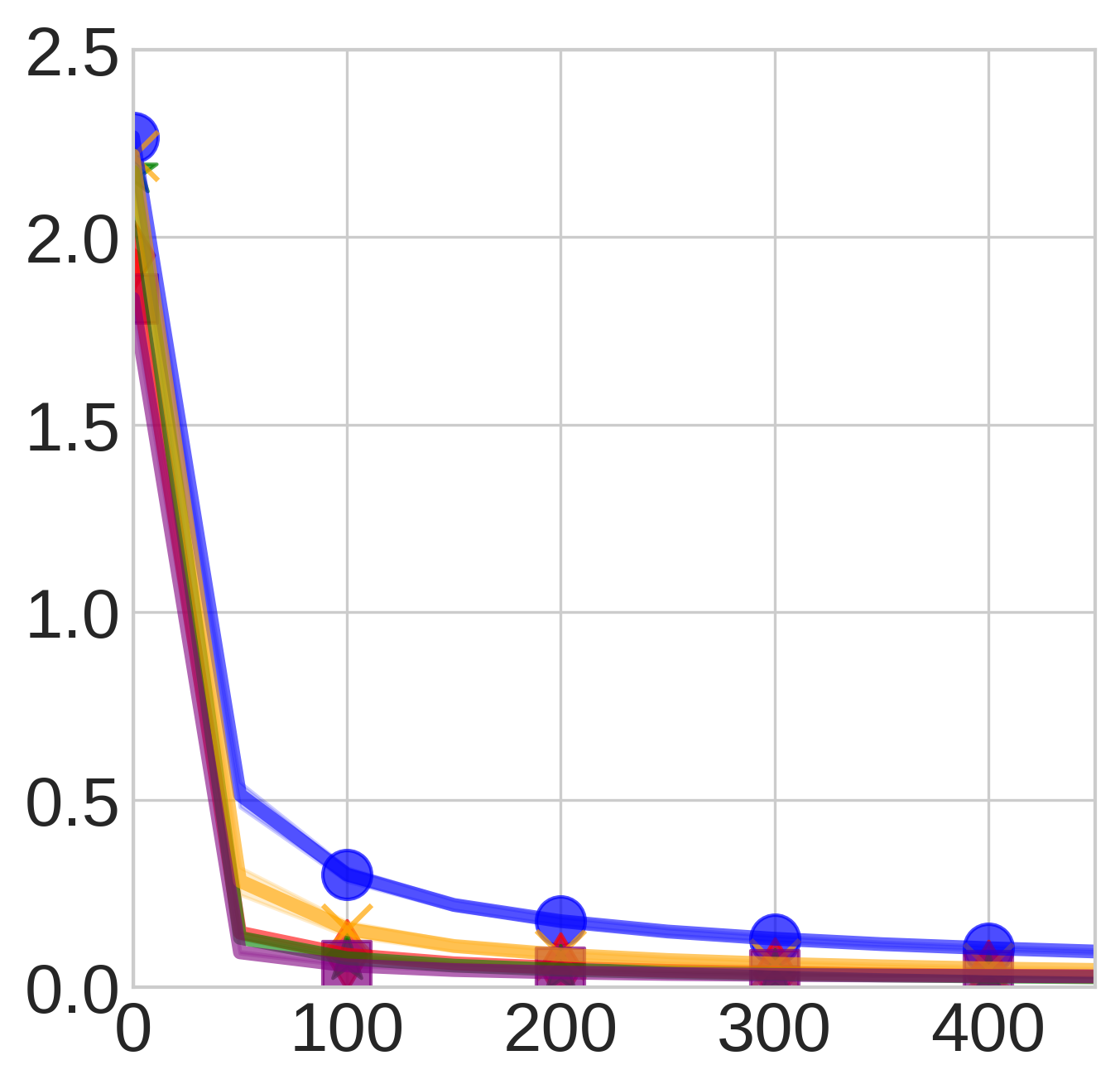}
	\label{}
\end{subfigure}\vspace{-0.15cm}
\begin{subfigure}[b]{0.15\textwidth}
	\centering
	\includegraphics[width=\textwidth]{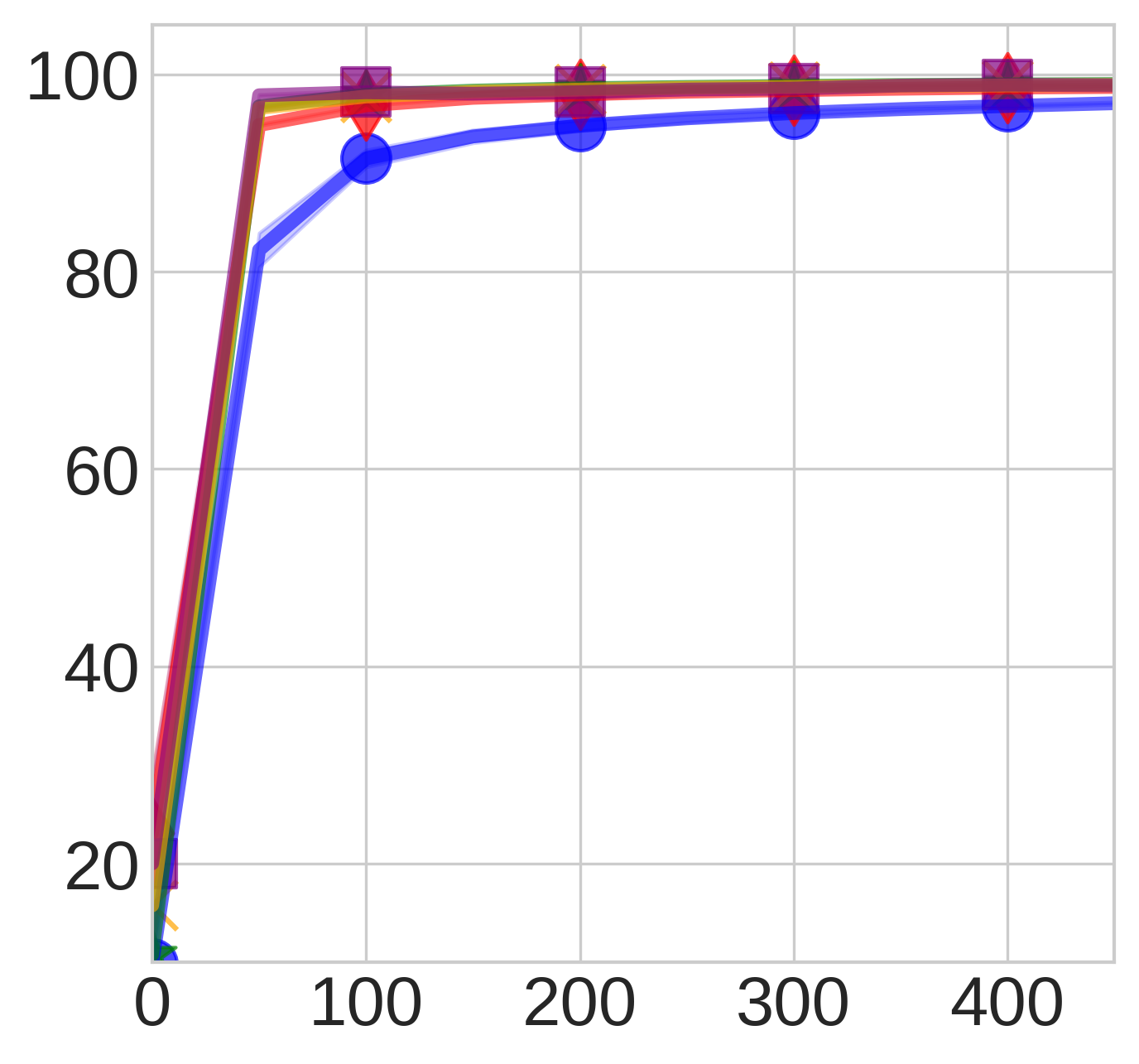}
	\label{}
\end{subfigure}\vspace{-0.15cm}
	\begin{subfigure}[b]{0.15\textwidth}
	\centering
	\includegraphics[width=\textwidth]{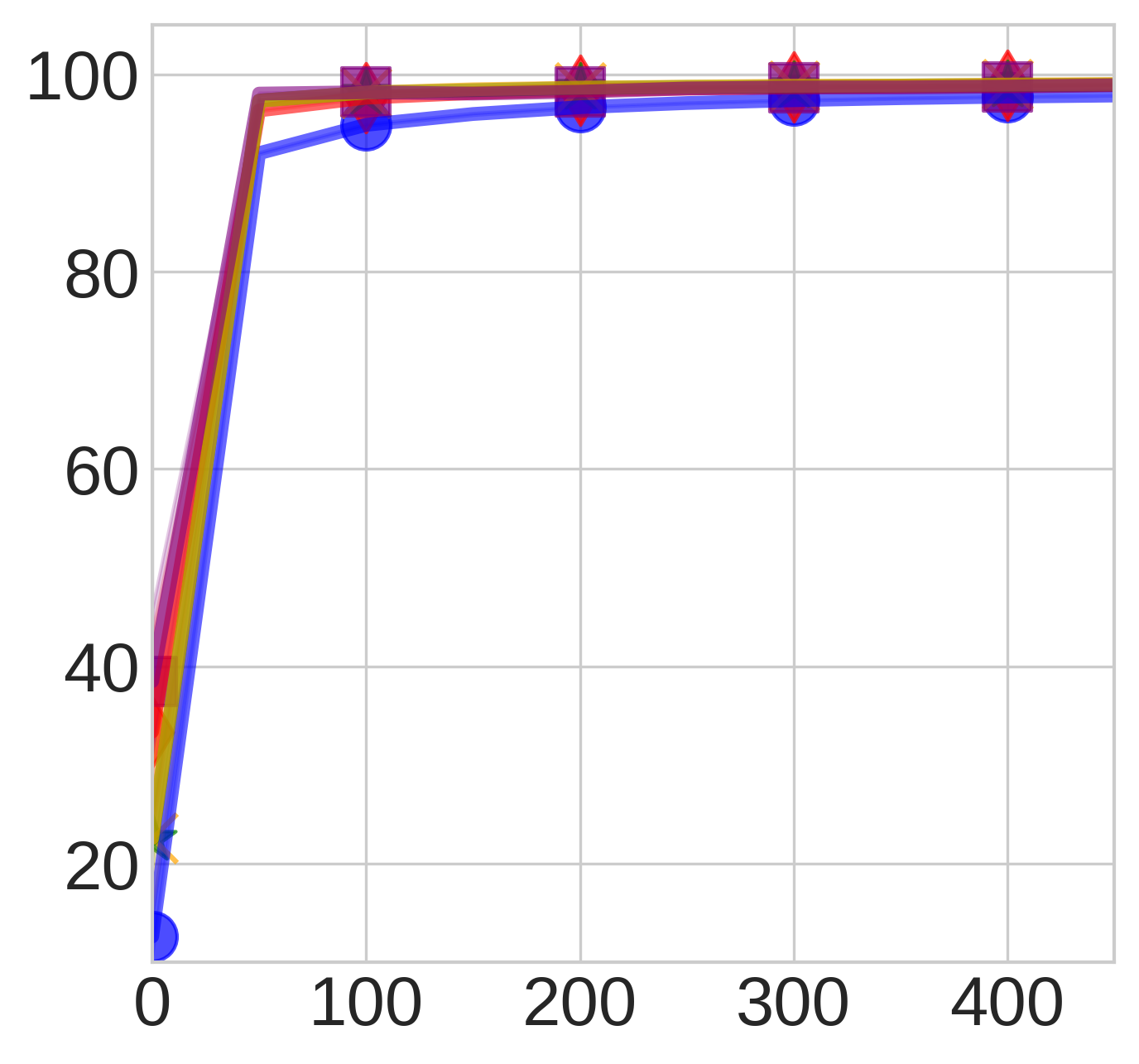}
	\label{}
\end{subfigure}\vspace{-0.15cm}
\begin{subfigure}[b]{0.15\textwidth}
	\centering
	\includegraphics[width=\textwidth]{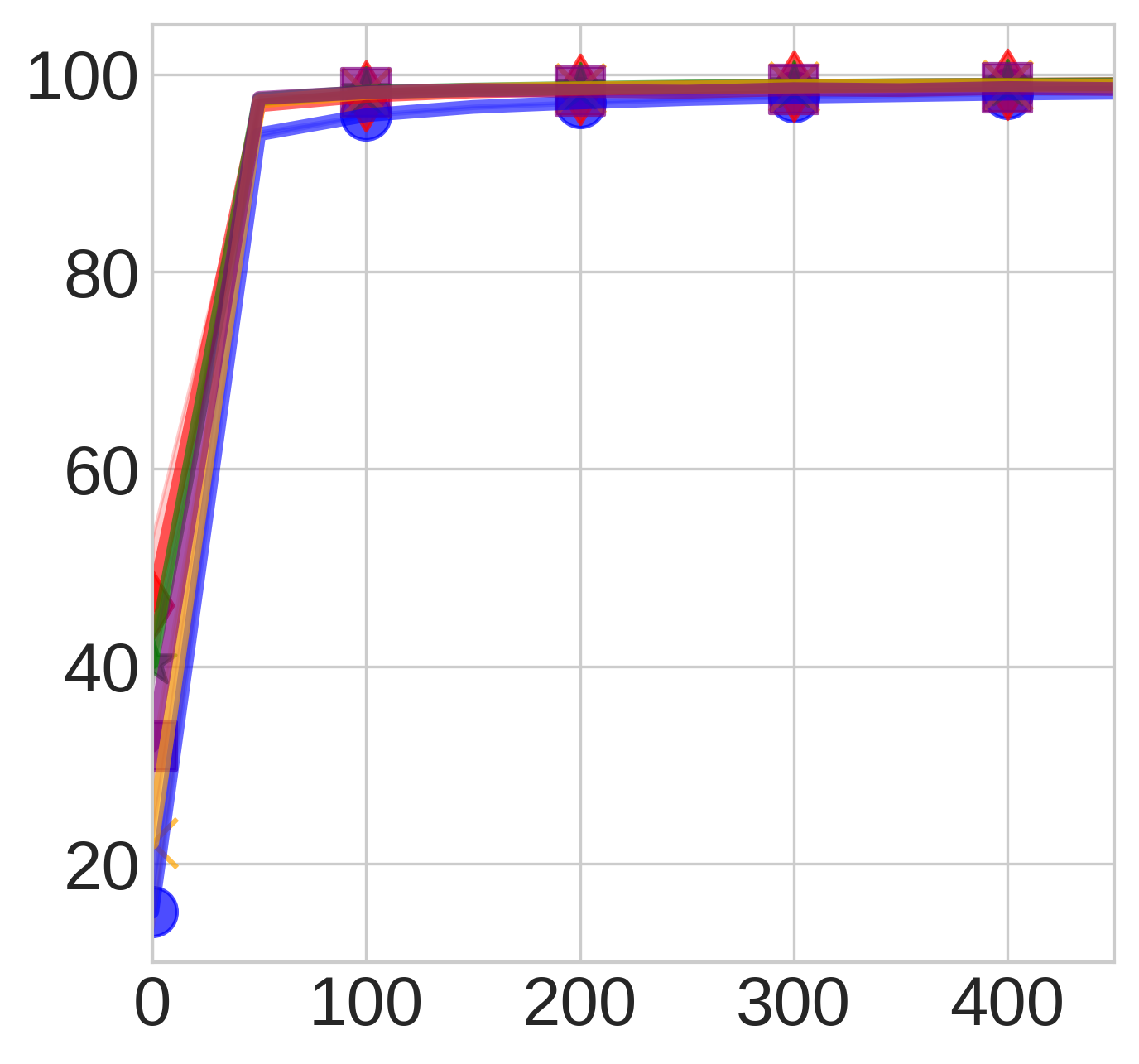}
	\label{}
\end{subfigure}\vspace{-0.15cm}
\begin{subfigure}[b]{0.15\textwidth}
    \centering
    \includegraphics[width=\textwidth]{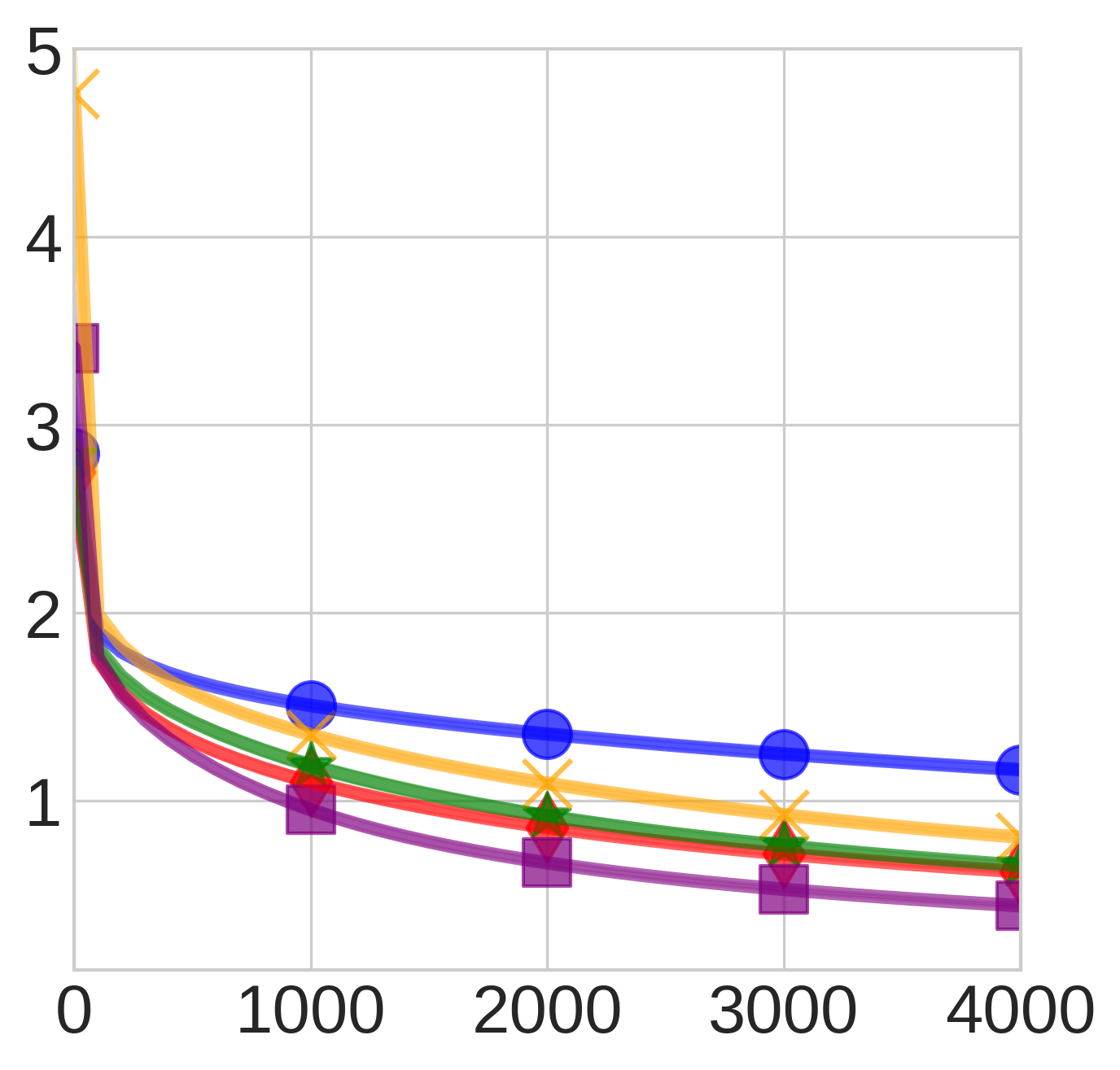}
    \label{}
  \end{subfigure}\vspace{-0.15cm}
  \begin{subfigure}[b]{0.15\textwidth}
    \centering
    \includegraphics[width=\textwidth]{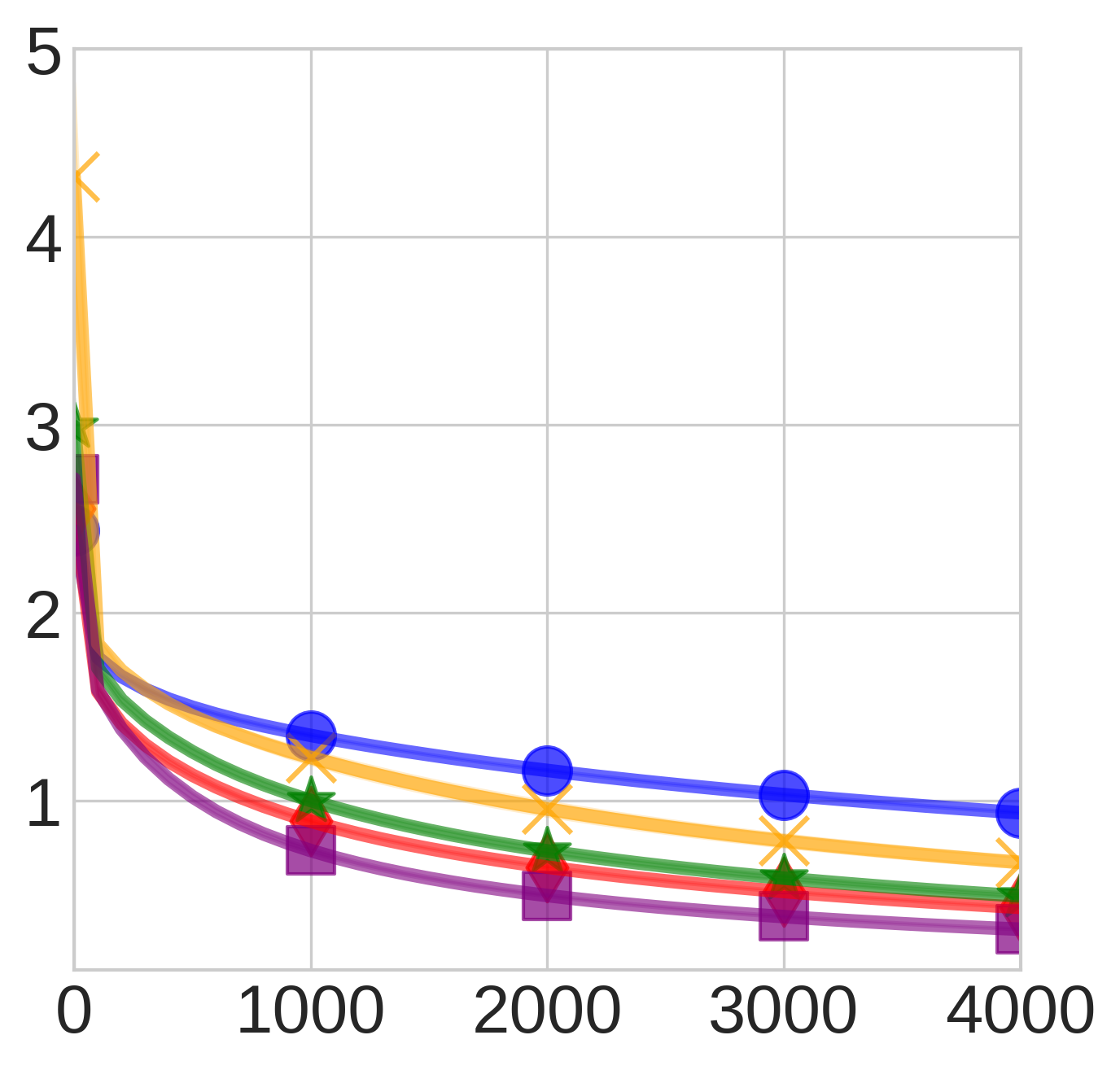}
    \label{}
  \end{subfigure}\vspace{-0.15cm}
  \begin{subfigure}[b]{0.15\textwidth}
    \centering
    \includegraphics[width=\textwidth]{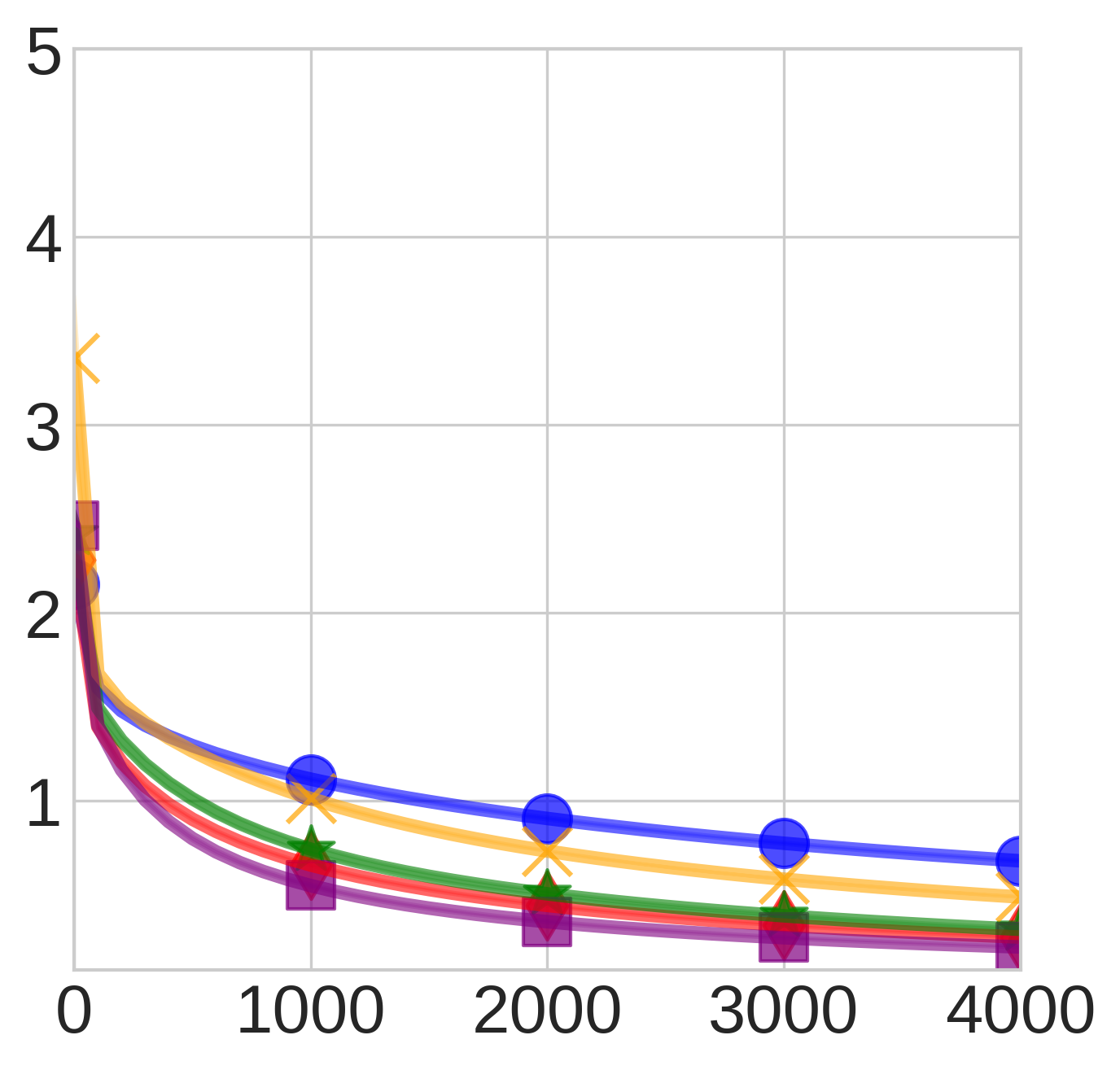}
    \label{}
  \end{subfigure}\vspace{-0.15cm}
  \begin{subfigure}[b]{0.15\textwidth}
    \centering
    \includegraphics[width=\textwidth]{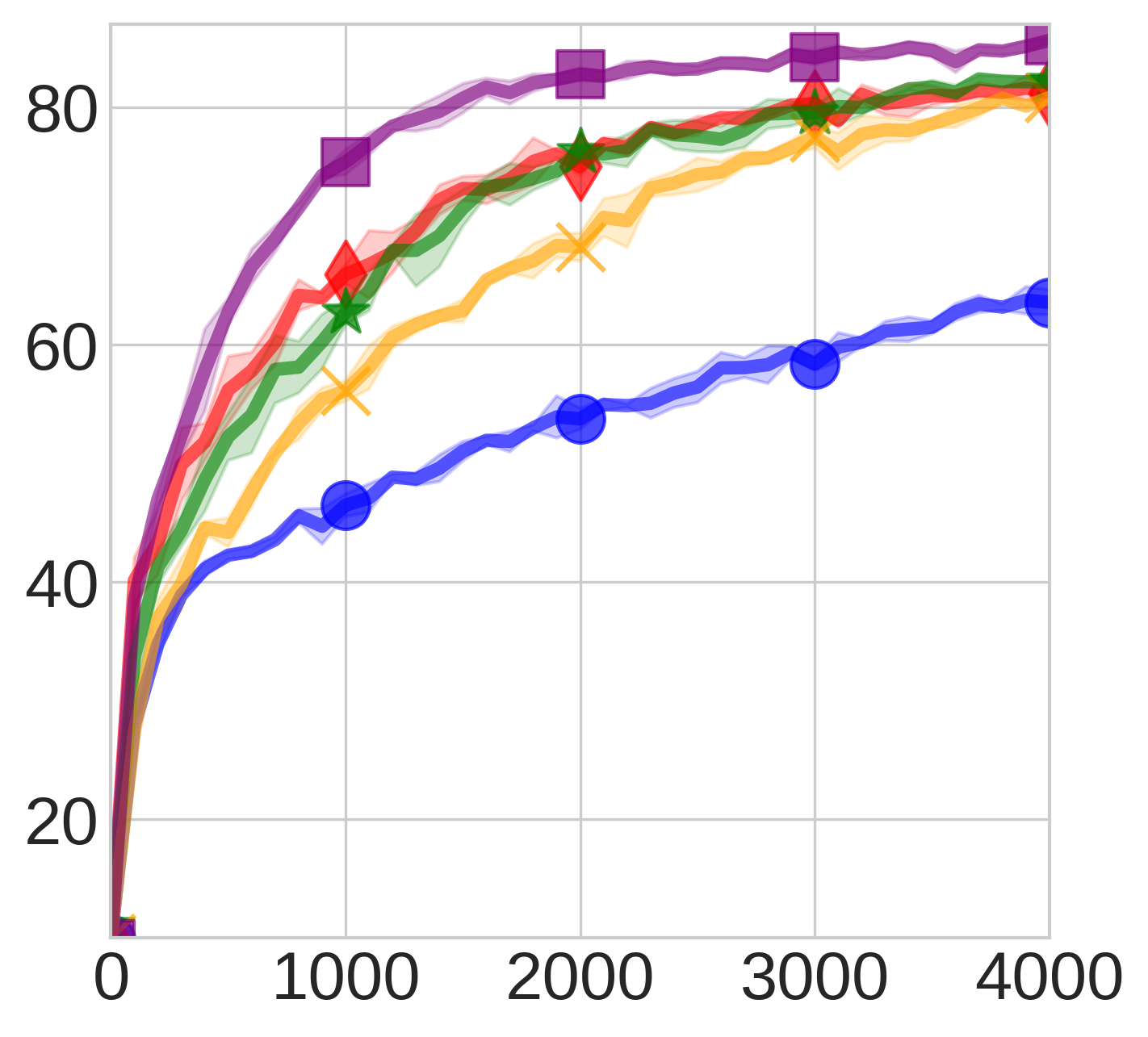}
    \label{}
  \end{subfigure}\vspace{-0.15cm}
    \begin{subfigure}[b]{0.15\textwidth}
    \centering
    \includegraphics[width=\textwidth]{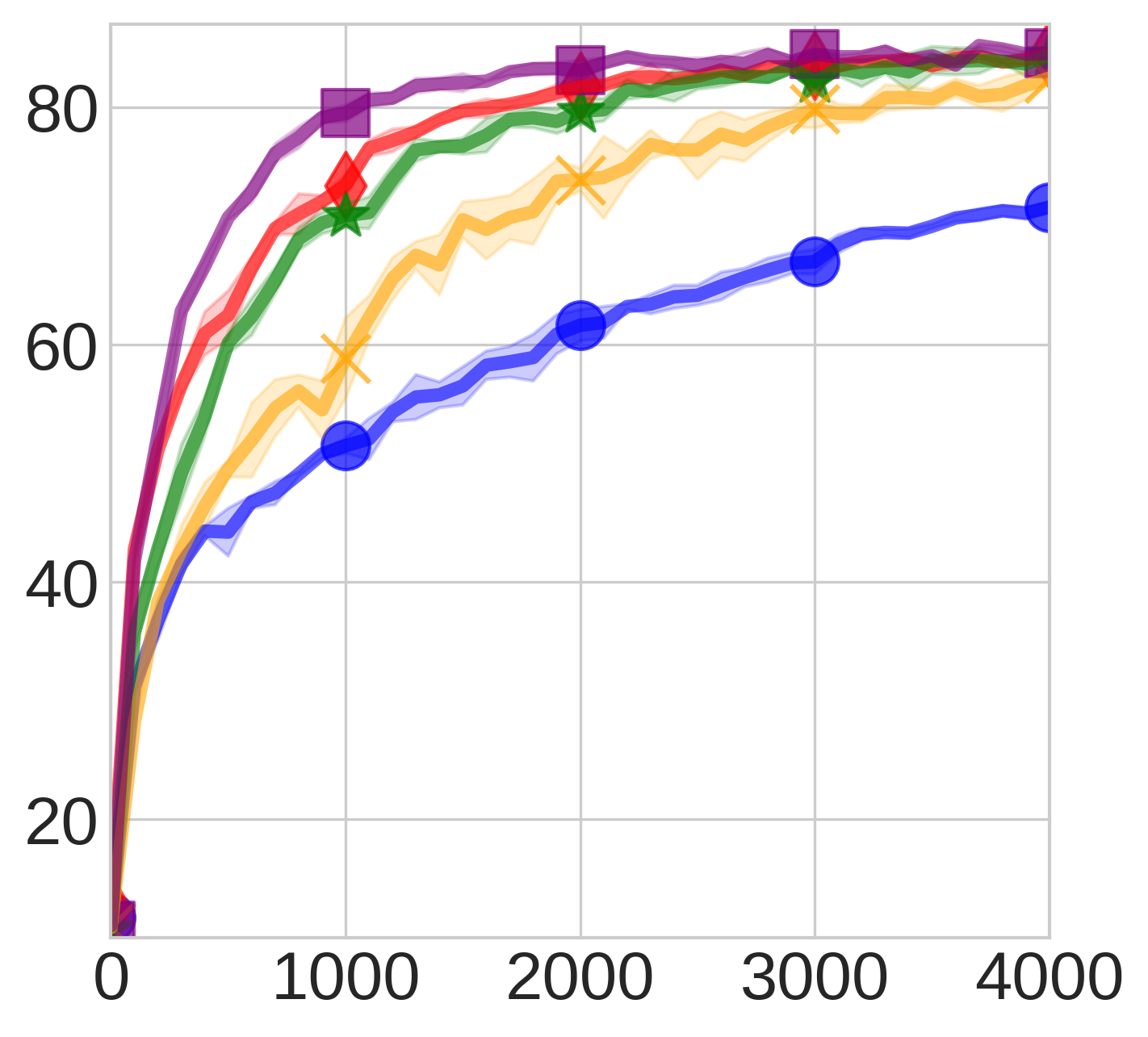}
    \label{}
  \end{subfigure}\vspace{-0.15cm}
  \begin{subfigure}[b]{0.15\textwidth}
    \centering
    \includegraphics[width=\textwidth]{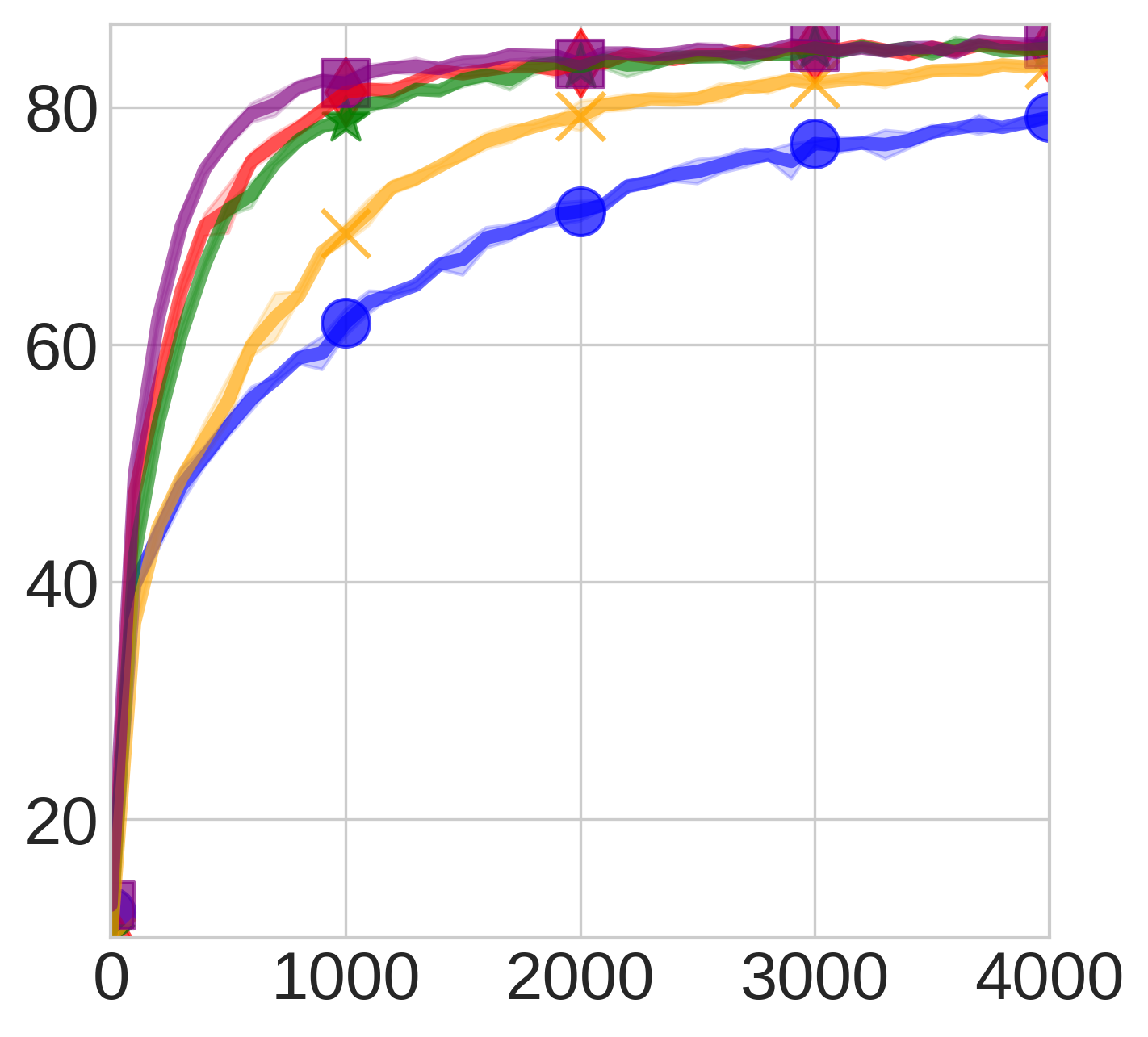}
    \label{}
  \end{subfigure}\vspace{-0.3cm}
\caption{Experimental results on the EMNIST and CIFAR-10 datasets. The initial two rows represent the outcomes obtained from the EMNIST, while the subsequent rows are the  results for CIFAR-10. The odd rows illustrate the training loss curves, while even rows depict the curves for test accuracy. For all the figures, the $x$-axis denotes the communication rounds. The columns, arranged from left to right, display the results obtained with varying values of $E=\{5, 10, 20\}$. } 
\label{fig:exp1}\vspace{-0.5cm}
\end{figure}

\vspace{-0.1cm} 
We summarize all experimental results in Fig. \ref{fig:exp1}. First of all, as we increase the parameter $E$ from $5$ to $20$, all algorithms exhibit accelerated performance due to the inclusion of more local steps. Besides, it is important to highlight that FedDA \cite{in2022accelerated} and FAFED \cite{wu2023faster}, both recognized as current state-of-the-art algorithms, have not previously been compared in the existing literature. Our findings from Fig. \ref{fig:exp1} reveal that the latest algorithm  FAFED \cite{wu2023faster} and MFL \cite{liu2020accelerating} surpass FedDA, while all adaptive algorithms consistently outperform FedAvg, underscoring the clear advantages of adaptive optimization techniques. Most notably, our proposed FedLion consistently outperforms all existing adaptive algorithms, thus establishing itself as the new state-of-the-art in adaptive algorithms across the two datasets under examination.

\vspace{-0.2cm}
\subsection{Gradient density}
\label{sec:grad_dense}
\begin{figure}[htbp]
	\centering
	\begin{subfigure}[b]{0.22\textwidth}
	\centering
	\includegraphics[width=\textwidth]{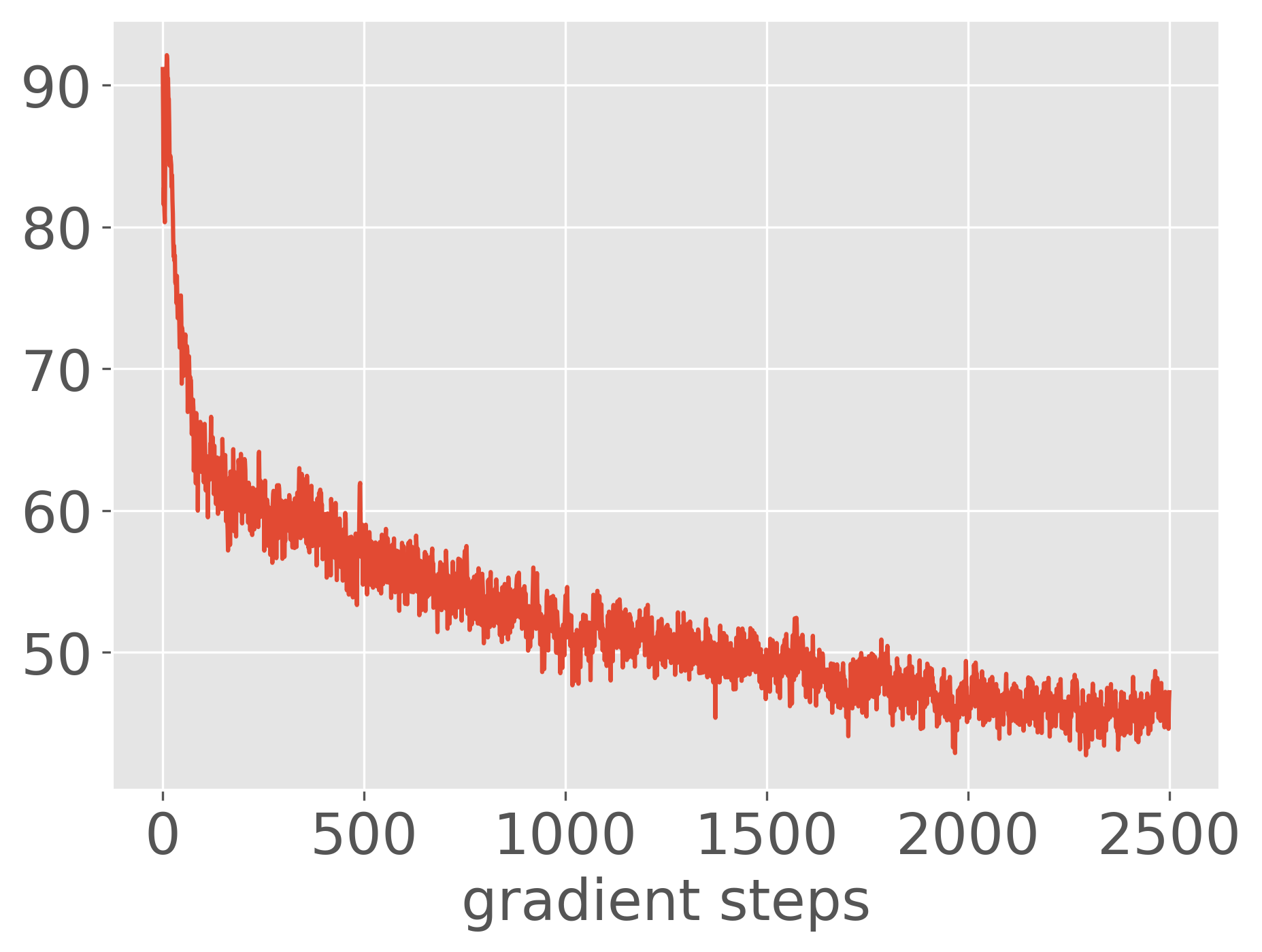}
	\caption*{}
	\label{}
\end{subfigure}
\begin{subfigure}[b]{0.23\textwidth}
	\centering
	\includegraphics[width=\textwidth]{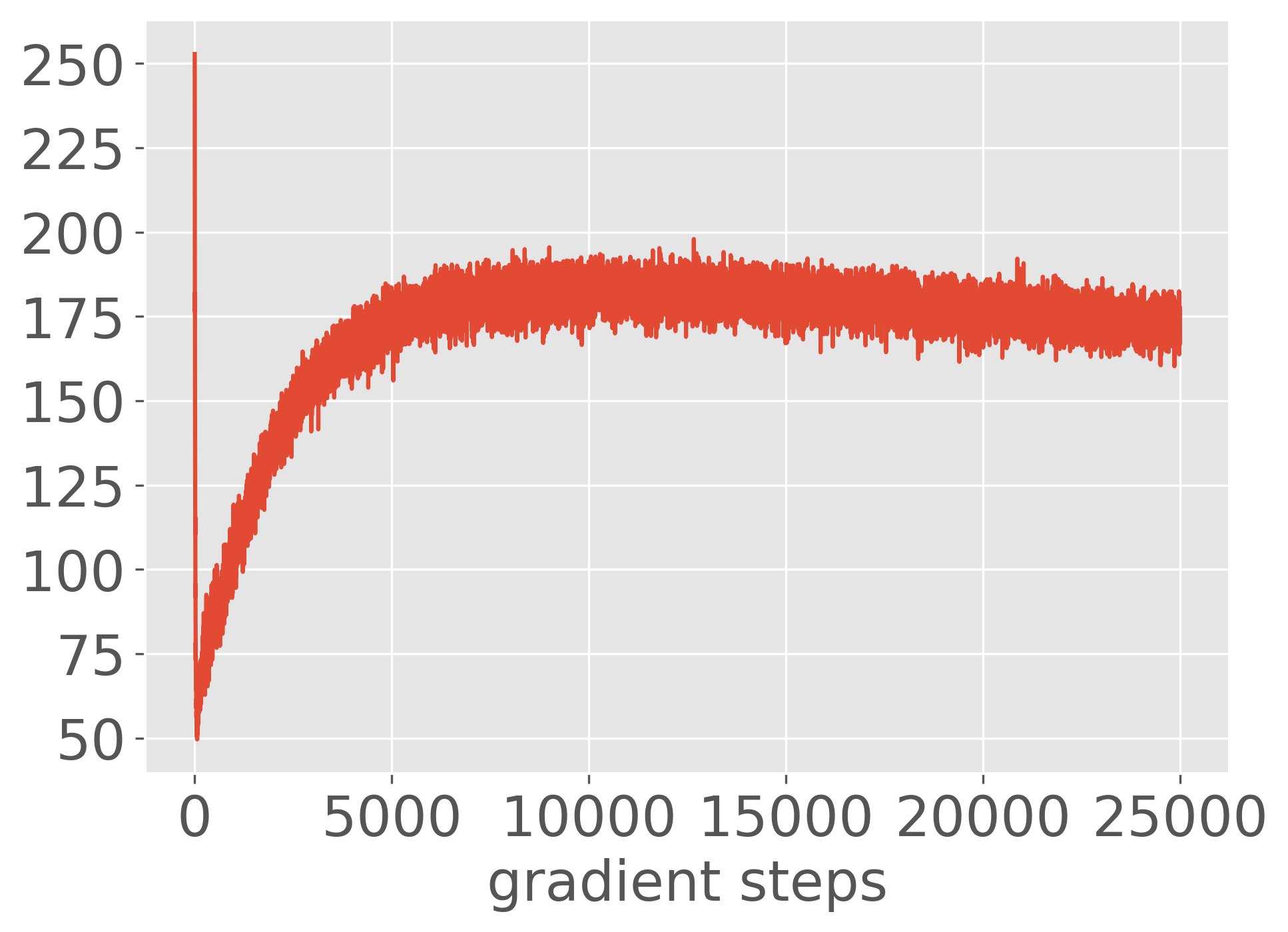}
	\caption*{}
	\label{}
\end{subfigure}
\vspace{-0.8cm}
\caption{Gradient density ${\|\tilde v_{t,s}\|_1}/{\|\tilde v_{t,s}\|_2}$ during training. Left is EMNIST and right is CIFAR-10. The curve is obtained by running FedLion with $E=5$.}
\label{fig:exp4}\vspace{-0.8cm}
\end{figure}

As previously discussed, our proposed algorithm demonstrates faster convergence when dealing with dense gradients, and can outperform existing state-of-the-art algorithms that primarily focus on $\ell_2$ norm convergence. In this section, we aim to substantiate our empirical observations regarding gradient density during the practical training of neural networks. To this end, we introduce two variables, $\tilde v_{t,s}=\frac{1}{n}\sum_{i=1}^n g_{t-1,s}^{i}$ and $v_{t,s}= \nabla f(\bar x_{t-1,s-1})$. In Fig. \ref{fig:exp4}, we provide empirical evidence by examining the density ${\|\tilde v_{t,s}\|_1}/{\|\tilde v_{t,s}\|_2}$ of the minibatch gradient during training, which serves as estimators for ${\|v_{t,s}\|_1}/{\|v_{t,s}\|_2}$. Fig. \ref{fig:exp4} clearly illustrates that in practical scenarios, the gradient tends to be dense (greater than $n^\frac{1}{4}$), thereby providing empirical justification for the observed accelerated convergence of our proposed algorithm.

\subsection{Intrinsic  quantization for uplink communication}
\begin{figure}[htbp]
	\centering
	\begin{subfigure}[b]{0.15\textwidth}
	\centering
	\includegraphics[width=\textwidth]{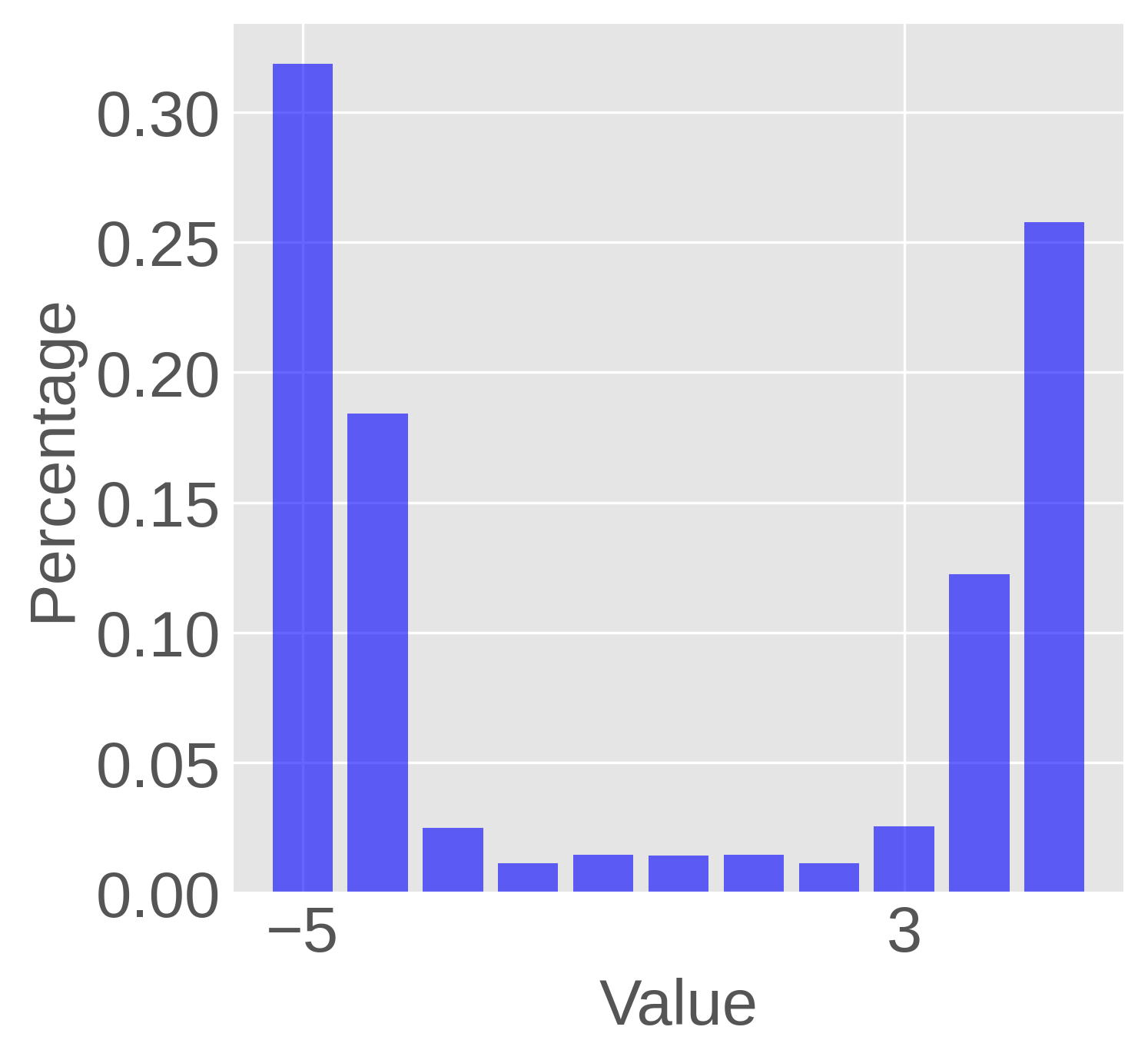}
	\caption*{}
	\label{}
\end{subfigure}
\vspace{-0.1cm}
\begin{subfigure}[b]{0.15\textwidth}
	\centering
	\includegraphics[width=\textwidth]{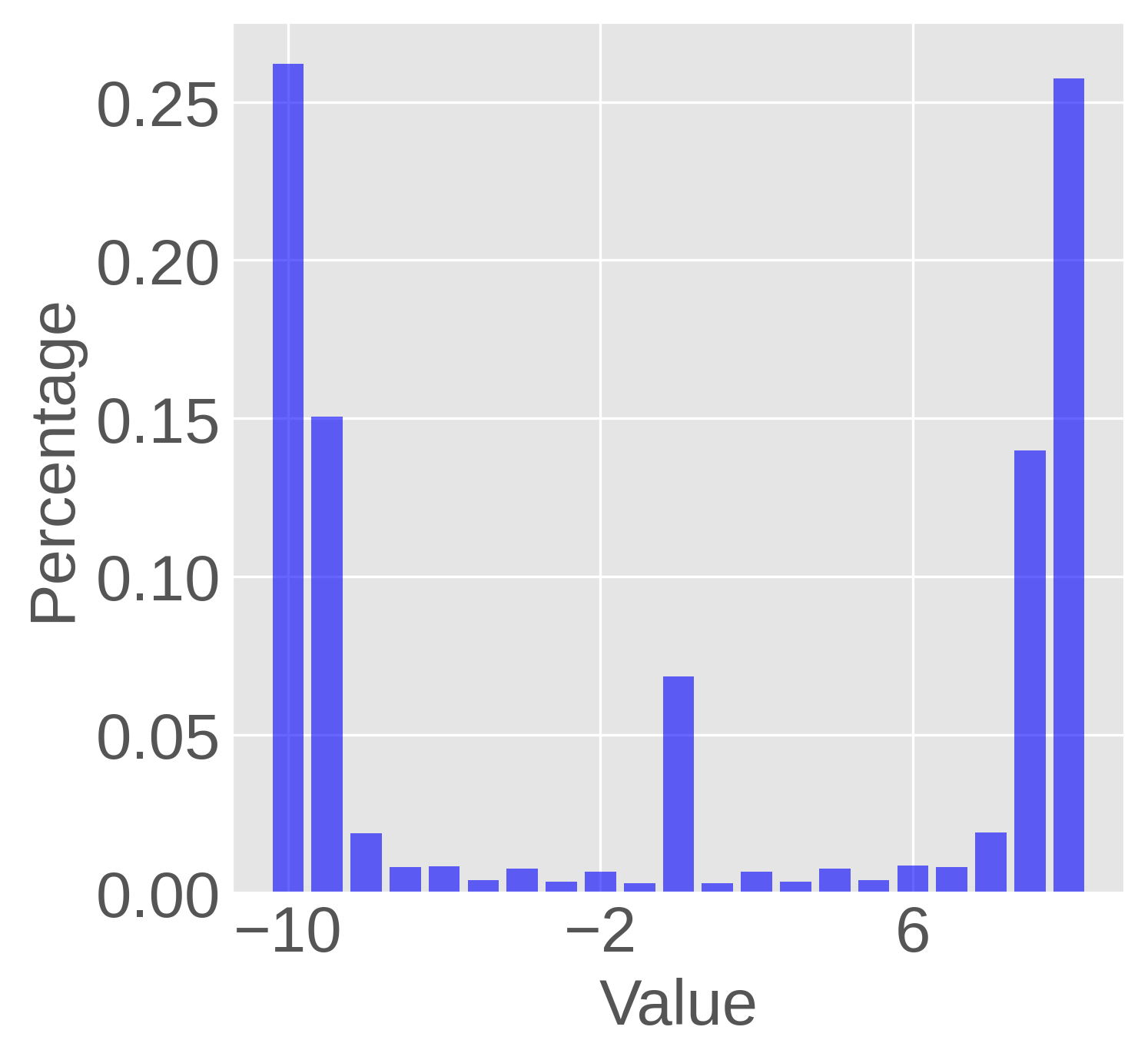}
	\caption*{}
	\label{}
\end{subfigure}
\vspace{-0.1cm}
\begin{subfigure}[b]{0.15\textwidth}
	\centering
	\includegraphics[width=\textwidth]{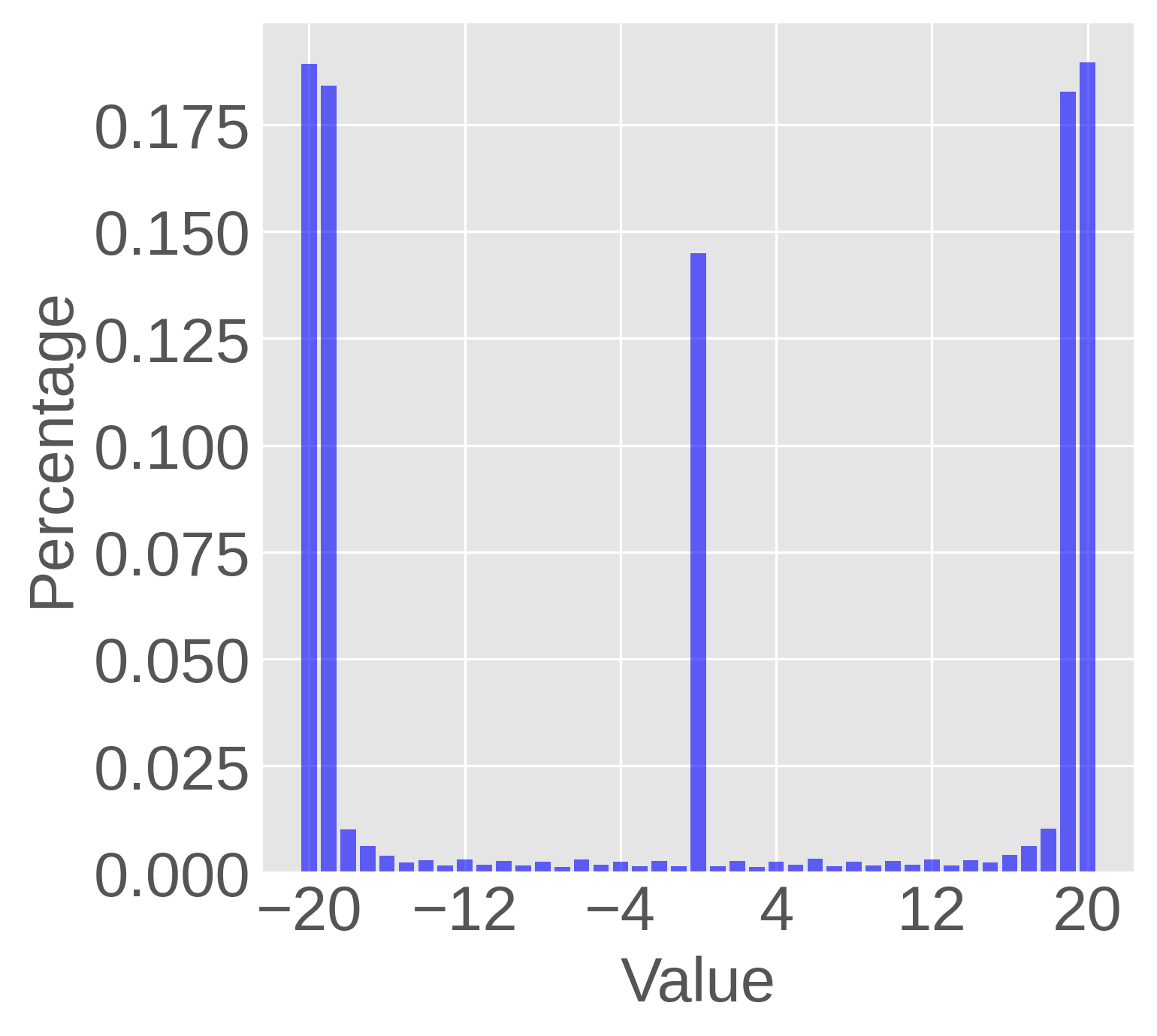}
	\caption*{}
	\label{}
\end{subfigure}
\vspace{-0.1cm}
\begin{subfigure}[b]{0.15\textwidth}
	\centering
	\includegraphics[width=\textwidth]{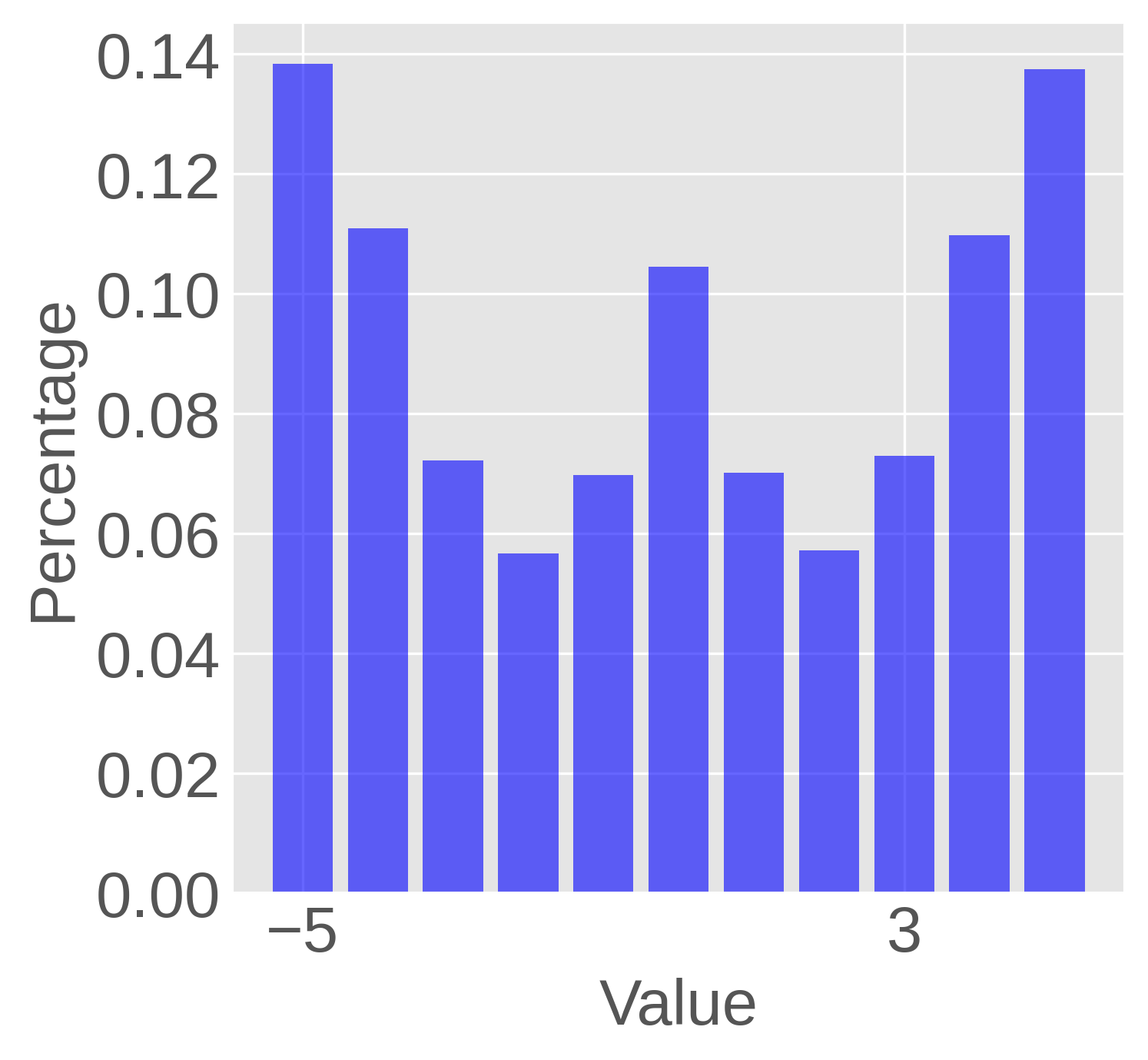}
	\caption*{}
	\label{}
\end{subfigure}
\vspace{-0.1cm}
\begin{subfigure}[b]{0.15\textwidth}
	\centering
	\includegraphics[width=\textwidth]{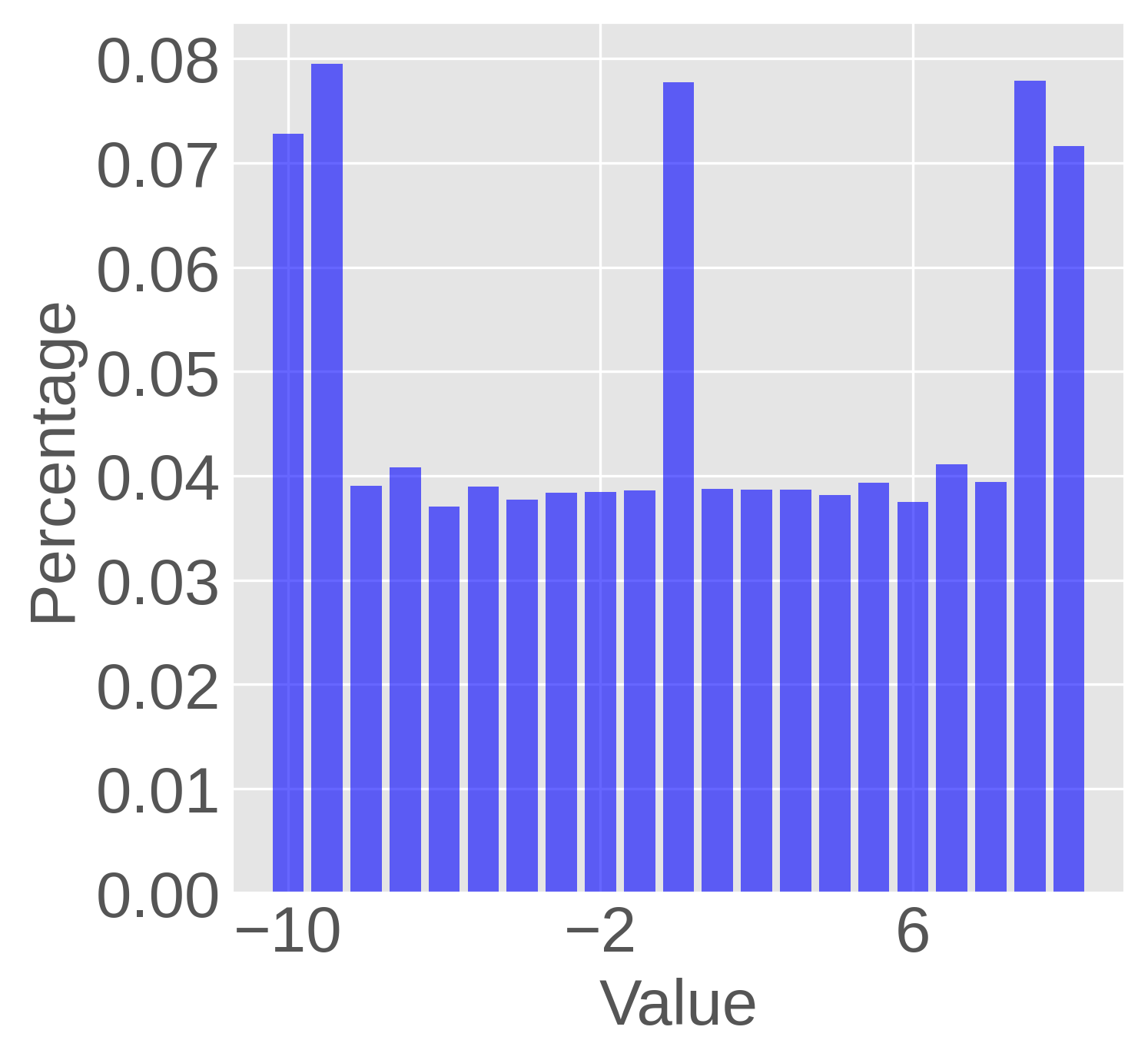}
	\caption*{}
	\label{}
\end{subfigure}
\vspace{-0.1cm}
\begin{subfigure}[b]{0.15\textwidth}
	\centering
	\includegraphics[width=\textwidth]{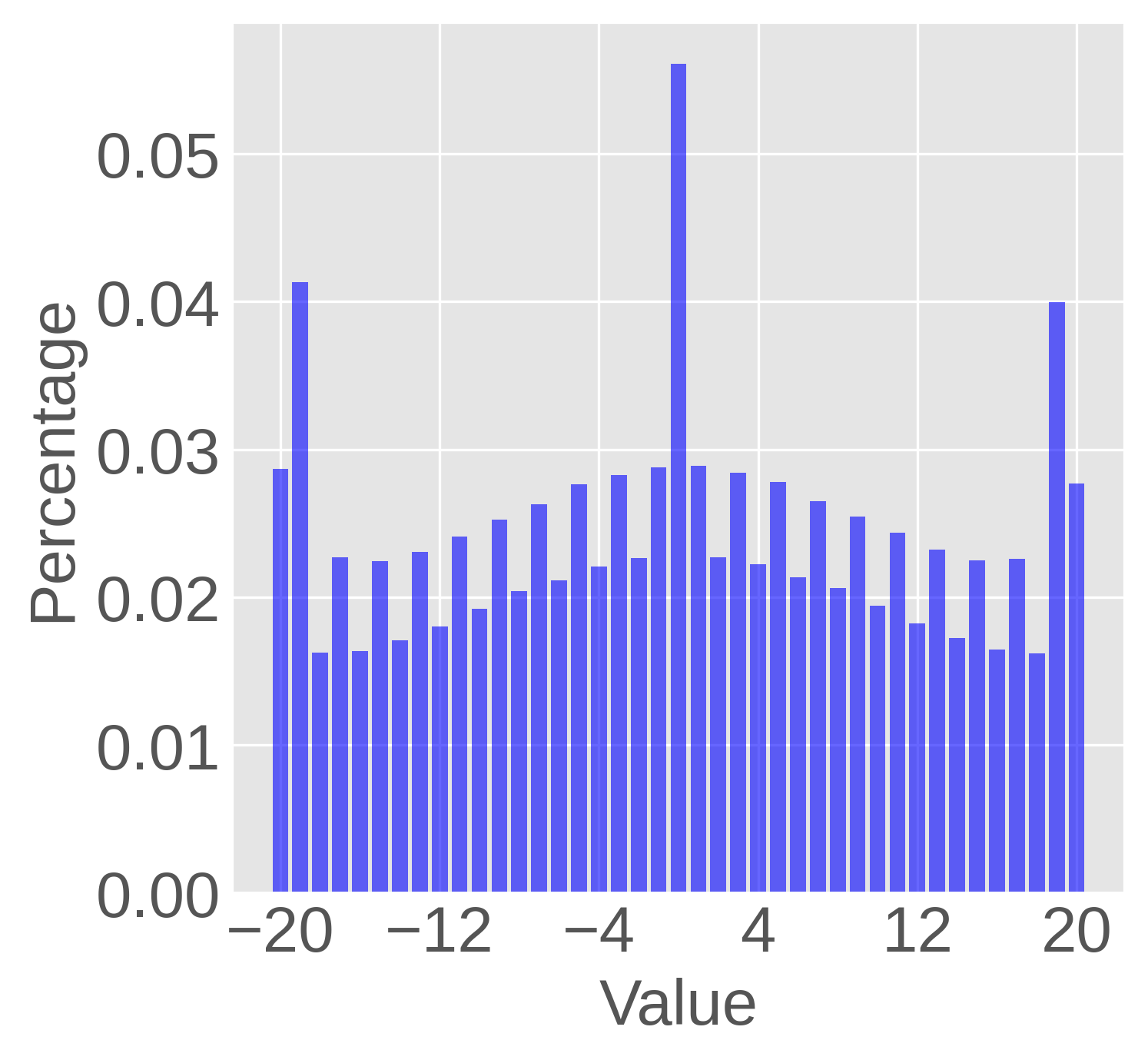}
	\caption*{}
	\label{}
\end{subfigure}
\vspace{-0.6cm}
\caption{Empiricial distribution of the element values in $\Delta_{t-1}^i$ during training. The first rows is on EMNIST dataset while the second rows is on CIFAR-10 dataset. From left tor right are the setting with $E\in\{5,10,20\}$ respectively.}
\label{fig:exp3} \vspace{-0.5cm} 
\end{figure}

\vspace{-0.2cm}
As explained in Section \ref{sec:algo}, we find that each element of $\Delta_{t-1}^{i}$ during uplink communication can be represented using a maximum of $\log(2E+1)$ bits. To delve deeper into this property, we conducted empirical investigations on the distribution of $\Delta_{t-1}^{i}$ values, considering various settings with $E\in\{5,10,20\}$ and present the results in Fig. \ref{fig:exp3}. In the case of the EMNIST dataset, we observe that values tend to cluster around the lower and middle ranges. This suggests that with efficient coding techniques, it's feasible to represent $\Delta_{t-1}^{i}$ using notably fewer bits than the theoretical requirement of $\log(2E+1)$ in practical scenarios. A similar trend is observed in the CIFAR-10 dataset, though with a more even distribution that might demand a higher bit count compared to the EMNIST dataset.

\vspace{-0.1cm} 
\section{Conclusion}
\label{sec:con} 
This work presents FedLion, a novel adaptive federated optimization algorithm that adapts Lion \cite{chen2023symbolic}, a state-of-the-art centralized algorithm, to federated learning. FedLion achieves superior theoretical convergence rates, particularly in scenarios with dense gradients. Empirical evaluations demonstrate the rapid convergence of FedLion, outperforming previous state-of-the-art algorithms such as FAFED \cite{wu2023faster} and FedDA \cite{in2022accelerated}, all while incurring only a minor increase in uplink transmission bits compared to FedAvg.

\section{Acknowledgments}
The work is supported by Shenzhen Science and Technology Program under Grant No. RCJC20210609104448114, the NSFC, China, under Grant 62071409, and by Guangdong Provincial Key Laboratory of Big Data Computing.

\bibliographystyle{IEEEbib}
\bibliography{template}
\nocite{*}
\end{document}